\documentclass[letterpaper, 10 pt, conference]{ieeeconf}  
\IEEEoverridecommandlockouts                              
\usepackage{amsmath,amsfonts,amssymb}
\usepackage{float}
\usepackage[mathscr]{euscript}
\usepackage{algorithm,algorithmicx}
\usepackage[noend]{algpseudocode}
\usepackage[dvipsnames]{xcolor}
\usepackage{graphicx}
\usepackage[font={small}]{caption}
\usepackage{subcaption}

\newcommand{\gbf}{\mathbf{g}}

\newcommand{\ellbf}{\mathbf{\ell}}

\newcommand{\sbf}{\mathbf{s}}

\newcommand{\xbf}{\mathbf{x}}

\newcommand{\zbf}{\mathbf{z}}

\newcommand{\Ebf}{\mathbf{E}}

\newcommand{\Ibf}{\mathbf{I}}

\newcommand{\Lbf}{\mathbf{L}}

\newcommand{\Pbf}{\mathbf{P}}

\newcommand{\Lcal}{\mathcal{L}}

\newcommand{\Ncal}{\mathcal{N}}
\newcommand{\Ocal}{\mathcal{O}}

\newcommand{\Scal}{\mathcal{S}}

\newcommand{\etabf}{\boldsymbol{\eta}}
\newcommand{\thetabf}{\boldsymbol{\theta}}

\newcommand{\lambdabf}{\boldsymbol{\lambda}}

\newcommand\SE[1]{{\mathsf{SE}(#1)}}
\newcommand{\Rbb}{\mathbb{R}}
\newcommand{\vcslam}{\textsc{vc-slam}}

\title{\LARGE \bf
	Variational Filtering with Copula Models for SLAM 
}

\author{John D. Martin$^{*,1}$, Kevin Doherty$^{*,2}$, Caralyn Cyr$^{1}$, Brendan Englot$^{1}$, John Leonard$^{2}$ 
\thanks{
$^{*}$ Equal contribution,
$^{1}$ Stevens Institute of Technology, 
$^{2}$ Massachusetts Institute of Technology, 
Correspondence to
        {\tt\small jmarti3@stevens.edu}}%
}

\begin{document}

\maketitle
\thispagestyle{empty}
\pagestyle{empty}

\begin{abstract}
The ability to infer map variables and estimate pose is crucial to the operation of autonomous mobile robots. In most cases the shared dependency between these variables is modeled through a multivariate Gaussian distribution, but there are many situations where that assumption is unrealistic. Our paper shows how it is possible to relax this assumption and perform simultaneous localization and mapping (SLAM) with a larger class of distributions, whose multivariate dependency is represented with a copula model. We integrate the distribution model with copulas into a Sequential Monte Carlo estimator and show how unknown model parameters can be learned through gradient-based optimization. We demonstrate our approach is effective in settings where Gaussian assumptions are clearly violated, such as environments with uncertain data association and nonlinear transition models. 
\end{abstract}

\section{Introduction}
Simultaneous localization and mapping (SLAM) \cite{smith1990estimating} is considered an essential capability for autonomous robots that operate in settings where precise maps and positioning are unavailable. SLAM enables robots to estimate their pose and landmark locations using a single inference procedure. Given generative models of how the latent pose evolves through time, and how observations relate to pose and landmarks, SLAM algorithms compute a posterior distribution and derive point estimates from its statistics. This paper studies a subclass of SLAM algorithms called \textit{filters}, used to make incremental posterior updates whenever new data becomes available.  

Conventional SLAM models impose several distributional and independence assumptions to keep inference tractable. One common assumption restricts poses and landmarks to be jointly Gaussian. Under a Gaussian model, multivariate dependency is represented as a linear function of covariances. Temporal dependency is often encoded with a Bayesian network or a factor graph \cite{slam:1993lu_milios:graph_slam, Thrun05, slam:kaess2010bayes} to sparsify the full joint Gaussian.  In spite of these sometimes unrealistic modeling choices, there are many algorithms that can accurately and efficiently recover the hidden variables \cite{slam:2007:grisetti_etal:rao_blackwell_pf, slam:2005:eustice_etal:sparse_delayed_pf, slam:2008:kaess_etal:isam}. 

Previous work has also pointed out ways in which these assumptions can be unrealistic \cite{Rosen13icra, slam:doherty2019multimodal, slam:fourie2016nonparametric}. Gaussian distributions are appropriate for real-valued variables whose noise is symmetric, unimodal, and lightly-tailed. However, this unimodality is not always appropriate when measurements and landmarks have an uncertain correspondence \cite{slam:doherty2019multimodal}. This can produce multiple hypotheses, or modes in the distribution, arising from several possible associations. Gaussian assumptions are also unrealistic when there exist complex relationships between the latent variables that cannot be modeled with a linear function of covariances. 
\begin{figure}
	\centering
	\includegraphics[width=\columnwidth]{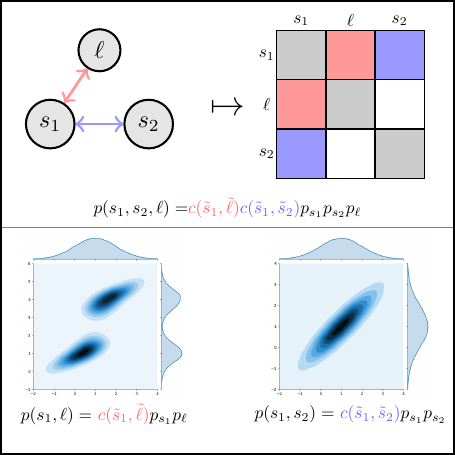}
	\caption{\textbf{Representing dependency for SLAM:} Graphical models are often employed to represent probabilistic dependence among latent variables and induce block structure in their (jointly Gaussian) information matrix (\textit{top half}). To move beyond Gaussian assumptions, we express the full joint distribution in terms of two dependency models (i.e. copulas), $c(\tilde{s}_1,\tilde{\ell}), c(\tilde{s}_1,\tilde{s}_2)$, and a product of all the marginals, $p_{s_1},p_{s_2},p_{\ell}$. Copulas represent the low-level multivariate dependence for relationships expressed in the graph topology. Copulas are agnostic to the distribution of marginal variables, permitting construction of arbitrary joint relationships our method exploits for improved SLAM inference (\textit{lower half}).}
	\label{fig:summary}
	\vspace{-1em}
\end{figure}

It stands to reason that by relaxing these assumptions, further improvements to accuracy could be realized. This is the main hypothesis our paper investigates; we introduce a family of factorized distributions whose dependency is modeled with a separate parametric function (Fig. \ref{fig:summary}) and argue that performing inference over this model class leads to improved accuracy and better representation of uncertainty.  

The models we introduce exploit Sklar's theorem \cite{prob:1959:sklar:copulas}, a foundational result from probability theory. Sklar's theorem states that \textit{any} joint distribution with a density, $p(X_1,\cdots,X_N)$, can be expressed in terms of a copula, $c: [0,1]^N \rightarrow [0,1]$, and a product of the associated marginal distributions: 
\begin{align}\label{eq:copula}
	p(X_1,\cdots,X_N) = c(\tilde{x}_1,\cdots,\tilde{x}_N)\prod_{i=1}^N p_i(X_i).
\end{align}
Here $\tilde{x}_i=F_i(x_i)$ is a univariate variable, output from the marginal cumulative distribution function (CDF) $F_i(X_i)$. Equation \ref{eq:copula} tells us that a joint probability can be decoupled into $N\in\mathbb{N}$ independent marginals $p_i(X_i)$ and a copula, $c(\cdot)$, to model the shared multivariate dependencies. 

\begin{figure}
    \centering
    \includegraphics[width=0.3\linewidth]{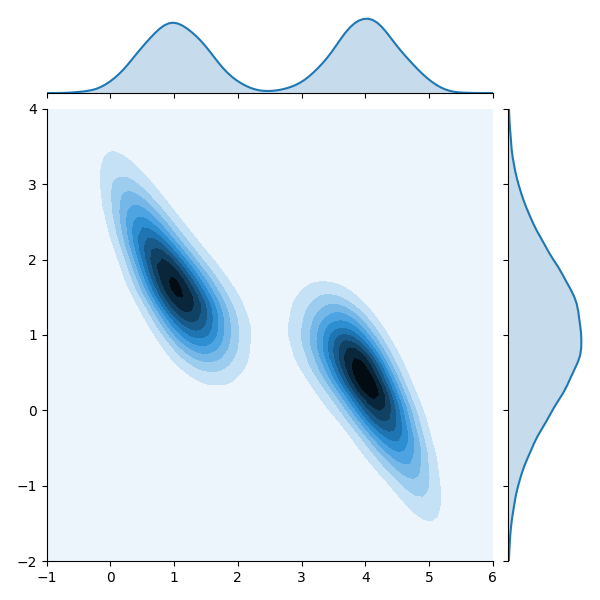}
    \includegraphics[width=0.3\linewidth]{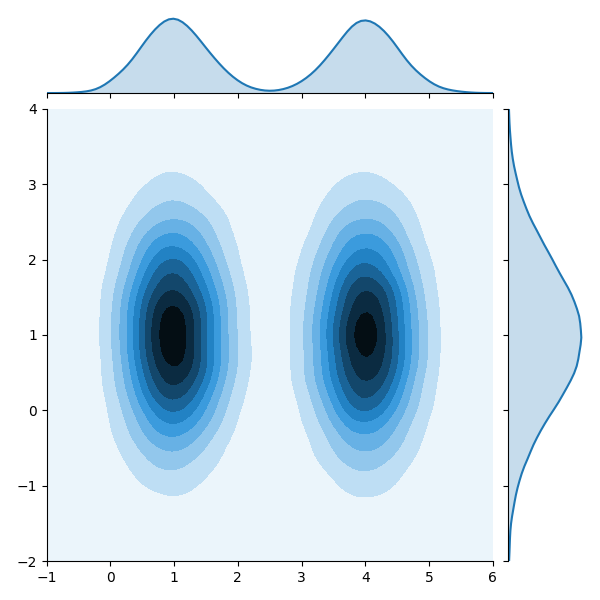}
    \includegraphics[width=0.3\linewidth]{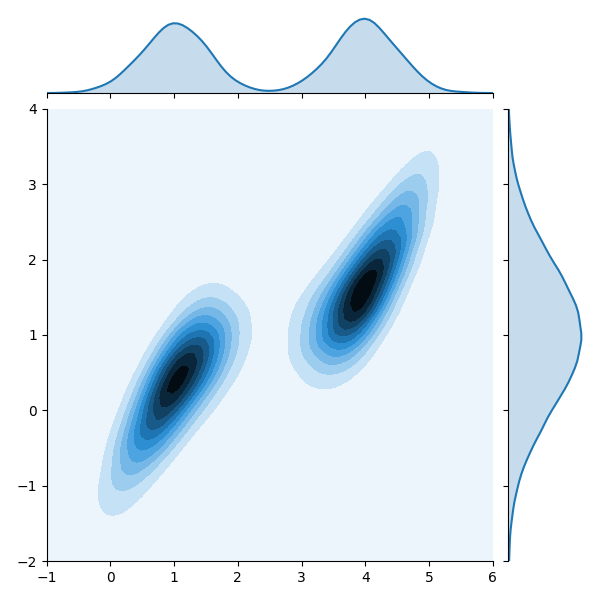}
    \caption{\textbf{Copulas can give rise to many distributions:} We show three distributions parameterized by different Gaussian copulas. From left to right, the correlation is $-0.9,0$, and $0.9$. The marginals are all the same: a two-component Gaussian mixture in $x$ and a unimodal Gaussian in $y$.}
    \label{fig:copula}
\end{figure}

Our paper shows how copulas provide a powerful mechanism to fuse together a set of any marginal distributions, and how they permit SLAM algorithms to reason about probabilistic dependence and marginal distributions separately (Fig. \ref{fig:copula}). The main contributions of our paper are as follows.

\textbf{A new class of models for SLAM:} We introduce a rich class of distributional models based on a copula factorization of the SLAM posterior. Models express a hierarchy of the joint relationships between latent variables, and they can capture non-Gaussian distributions while still remaining fully factorized for efficient inference. 


\textbf{A variational filter for non-Gaussian SLAM:} We introduce a new algorithm, called VC-SMC (Variational Copula Sequential Monte Carlo), that learns the parameters of a copula-factorized distribution model. VC-SMC is based on a particle filter. This sequentially generates data used to fit the variational parameters, making it possible to perform SLAM when the data has complex dependencies and the model parameters are unknown.

\section{Preliminaries}
\subsection{Simultaneous Localization And Mapping (SLAM)}
We consider a standard observation process for navigation: at each step $t\in\mathbb{N}$ a mobile robot in state\footnote{In general states can be from any vector space. This work considers $\Scal= \SE{2}$, for vehicles that operate in the two-dimensional plane.} $\sbf_t \in \Scal$ transitions to a state $\sbf_{t+1} \sim p(\cdot|\sbf_{1:t})$ and detects some set of $M_t\in\mathbb{N}$ landmarks\footnote{Each landmark $\ellbf$ is represented with a vector of spatial coordinates: two for planar navigation problems and three for 3D navigation.}, $\ellbf_{1:M_t}$, from the observation $\zbf_t\sim p(\cdot|\sbf_{1:t},\ellbf_{1:M_{1:t}})$. The goal is to infer the robot's history of states and the landmark locations given the full sequence of observations. We aggregate the landmarks and robot state into the single latent variable $\xbf_t = (\sbf_t,\ellbf_{1:M_t})$. The joint distribution of this process is given by $p(\xbf_{1:T},\zbf_{1:T}) =$
\begin{align}\label{eq:slam_joint}
 p(\xbf_1)p(\zbf_1|\xbf_1)\prod_{t=2}^Tp(\xbf_t|\xbf_{1:t-1})p(\zbf_t|\xbf_{1:t},\zbf_{1:t-1}). 
\end{align}
This general model captures many common state-space systems, Hidden Markov Models (HMM), and non-Markov models, such as Gaussian Processes \cite{gp:roger_etal:vgp_statespace} and those represented by recurrent neural networks \cite{smc:2015:gu_etal:neural_smc}.

\subsection{Sequential Monte Carlo}
\begin{figure}
	\centering
	\includegraphics[width=\columnwidth]{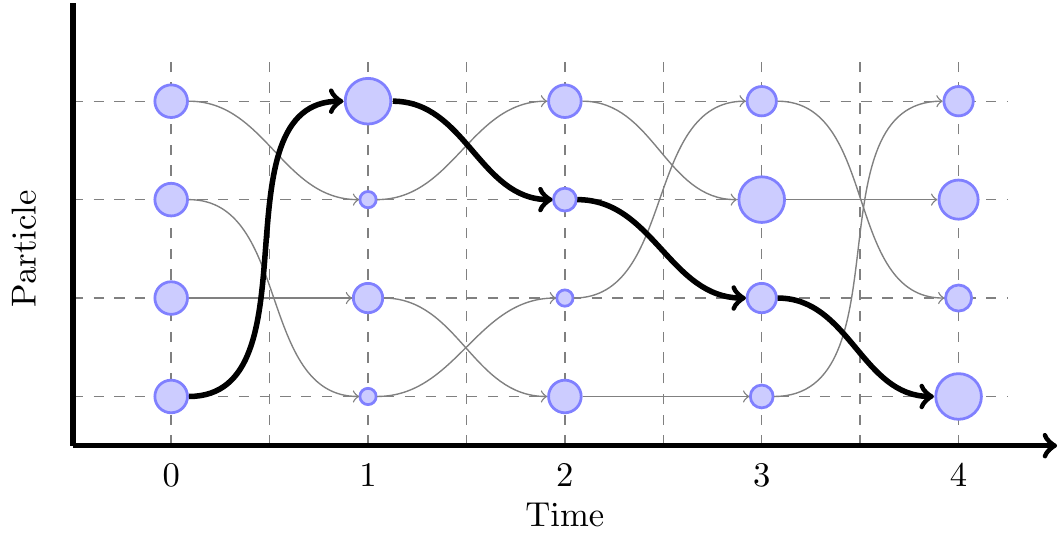}
	\caption{\textbf{Sequential Monte Carlo sampling:} These methods construct a weighted set of particles whose sequences form trajectories. Circle size is proportional to the weight $w_t$. Arrows denote ancestor links, and the black path denotes the chosen trajectory. Our procedure selects a trajectory based on its final weight size.}	
	\label{fig:smc}
\end{figure}

We consider Sequential Monte Carlo (SMC) methods to solve the SLAM problem (Fig. \ref{fig:smc}). SMC methods approximate the posterior with a weighted set of $N$ sampled trajectories $p(\xbf_{1:T}|\zbf_{1:T}) \approx \frac{1}{N}\sum_{n=1}^N w_T^{[n]}\delta(\xbf_{1:T}^{[n]})$, which are drawn from a simpler proposal distribution $\{\xbf^{[n]}_{1:T}\}_{n=1}^N\sim q(\xbf_{1:T}|\zbf_{1:T})$ \cite{smc:doucet2001introduction}. The joint distribution of latent variables and observations is given in \eqref{eq:slam_joint}. Although any distribution can be used for the proposal, it is beneficial to choose one with an autoregressive structure: 
\begin{align*}
q(\xbf_{1:T}|\zbf_{1:T}) = q(\xbf_{1}|\zbf_{1})\prod_{t=2}^T q(\xbf_{t}|\xbf_{1:t-1},\zbf_{1:t}).
\end{align*} 
This makes it possible to decompose the complete proposal into $T$ conditional distributions. The normalized importance weights of the $n$-th sample are defined as 
\begin{align*}
	w_t(\xbf_{1:t}^{[n]})= \frac{p(\xbf_{1:t}^{[n]}|\zbf_{1:t})}{q(\xbf_t^{[n]}|\xbf_{1:t-1}^{[n]},\zbf_{1:t})}.
\end{align*}
They are computed with the following approximation:
\begin{align}
	w_t(\xbf_{1:t}^{[n]})&\approx \frac{ \tilde{w}_t(\xbf_{1:T}^{[n]})}{\sum_{n=1}^N \tilde{w}_t(\xbf_{1:T}^{[n]})}, \label{eq:smc_weights}\\
	 \tilde{w}_t(\xbf_{1:t}^{[n]}) &= \tilde{w}_{t-1}(\xbf_{1:t-1}^{[n]})\cdot \xi(\xbf_{1:t}^{[n]},\zbf_{1:t}),\\
	 \xi(\xbf_{1:t}^{[n]},\zbf_{1:t}) &= \frac{ p(\xbf_t^{[n]}|\xbf_{1:t-1}^{[n]})p(\zbf_t|\xbf_{1:t}^{[n]},\zbf_{1:t-1})}{q(\xbf_{t}^{[n]}|\xbf_{1:t-1}^{[n]},\zbf_{1:t})}.
\end{align}
The weighted set of particles is constructed sequentially. At time $t=1$, we use standard importance sampling \cite{smc:doucet2001introduction}. Beyond that, each step begins by resampling ancestor index variables $a^{[n]}_{t-1}\in\{1,\cdots,N\}$ with probability proportional to the importance weights \eqref{eq:smc_weights}. Next, new values are proposed by drawing samples from $q(\xbf_{t}^{[n]}|\xbf_{1:t-1}^{a_{t-1}^{[n]}},\zbf_{1:t})$. The new samples are appended to the end of the trajectory, $\xbf_{1:t}^{[n]} = (\xbf_{1:t-1}^{a_{t-1}^{[n]}},\xbf_{t}^{[n]})$, and weights are updated with \eqref{eq:smc_weights}. 

The accuracy of SMC methods critically relies on the proposal distribution. One common choice, known as the \textit{Bootstrap Particle Filter (BPF)} \cite{gordon1993novel}, uses the transition model for the proposal: 
\begin{align}\label{eq:bootstrap}
q_{\textnormal{BPF}}(\xbf_t|\xbf_{1:t-1})=p(\xbf_t|\xbf_{1:t-1}).
\end{align}
The BPF has been employed in robotics settings \cite{slam:montemerlo2002fastslam}, and is known to be inaccurate in high dimensions and when there are few particles available to sample. To remedy this issue, we propose a learning procedure that adjusts parameters of the proposal distribution, which is a copula-factorized model. 

\subsection{Copula models}
We consider proposal distribution models whose multivariate dependency is modeled with a copula. Different copula models give rise to different types of multivariate relationships. One common model we use is the Gaussian copula, which is given in terms of the zero-mean multivariate CDF $\Phi_{\Sigma}(\cdot)$ and its inverse $\Phi_{\Sigma}^{-1}(\cdot)$:
\begin{align}\label{eq:gaussian_copula}
	C_{\Sigma}(\tilde{\xbf}_1,\cdots,\tilde{\xbf}_N) &= \Phi_{\Sigma}(\Phi^{-1}(\tilde{\xbf}_1),\cdots,\Phi^{-1}(\tilde{\xbf}_N)).
\end{align}
Here $\Sigma$ is the covariance matrix, and  $\tilde{\xbf}_i=\Phi_{\Sigma}(\xbf_i)$. Notice the transformed variables $\tilde{\xbf}_n$ are uniformly distributed, since CDFs map their inputs to the unit interval. Figure \ref{fig:copula} shows how the Gaussian copula can give rise to many complex distributions which are not Gaussian themselves. Other copula models that focus on two-dimensional relationships are the Student-t, Clayton, Gumbel, Frank, and Joe copulas \cite{prob:nelson:an_introduction_to_copulas}. Copula models can be applied in a hierarchy, and their parameters can be learned with variational inference \cite{vi:2015:tran:copula_variational_inference}.

\section{Copula-factorized Distributions for SLAM}
In this section we introduce a class of distributions for SLAM, aimed to model the joint relationship among latent variables with a hierarchy of copula models. We apply Sklar's theorem \eqref{eq:copula} to factor the full SLAM posterior into several copulas and a product of univariate marginals. The copula hierarchy we present is one of several plausible dependency structures that could be useful for estimation. We propose four time-invariant relationships intended to promote parameter sharing. Other structures could be chosen to trade off computational constraints for model accuracy. Throughout the section we denote univariate variables with superscript parentheses, $\xbf^{(i)}$, and the copula parameters relating $a$ and $b$  are $\thetabf(a,b)$. For instance, the parameters of a Gaussian copula are the elements of the correlation matrix. 

\paragraph{States and landmarks}The first relationship we consider is between the states and landmarks. We apply the decomposition of \eqref{eq:copula} by factoring these two variables and introducing a copula function $c_{\thetabf(s,\ell)}$ to model their joint relationship:
\begin{align*}
	p(\sbf_t,\ellbf_{1:M_t}|\zbf_{1:t}) &= c_{\thetabf(s,\ell)}(\tilde{\sbf}_t,\tilde{\ellbf}_{1:M_t})p(\sbf_t|\zbf_{1:t})p(\ellbf_{1:M_t}|\zbf_{1:t}).
\end{align*}
In scenarios where perturbations in either $\sbf$ or $\ellbf$ cause variations to the other, dependency should be represented in the model. In such cases, $c_{\thetabf(s,\ell)}$ stands as a means to model that relationship.
 
\paragraph{Landmark dependency}Relationships commonly exist among landmarks when their positions are influenced by the same phenomena. For instance, two landmarks may be a part of the same dynamic structure. When they do, their joint distribution can be modeled as
\begin{align*}
	p(\ellbf_{1:M_t}|\zbf_{1:t}) &=c_{\thetabf(\ell,\ell)}(\tilde{\ellbf}_{1:M_t})\prod_{j=1}^{M_t}p(\ellbf_{j}|\zbf_{1:t}).
\end{align*}
This assumes homogeneity among the landmarks' relationships, so that dependence can be modeled with a common copula $c_{\thetabf(\ell,\ell)}$. Without homogeneity, there could be ${M_t \choose 2}$ copulas in the worst case. 
\paragraph{Component dependency}Another common source of dependency exists between the scalar components of the landmarks $c_{\thetabf(\ell)}$ and states $c_{\thetabf(s)}$. These relationships represent more familiar dependency encoded in the transition and observation models. Each model factorizes as follows
\begin{align*}
	p(\ell_j|\zbf_{1:t}) &= c_{\thetabf(\ell)}(\tilde{\ell}^{(1:d_\ell)}_t)\prod_{i=1}^{d_\ell}p(\ellbf^{(i)}_t|\zbf_{1:t}),\\
	p(\sbf_t|\zbf_{1:t}) &= c_{\thetabf(s)}(\tilde{\sbf}^{(1:d_s)}_t)\prod_{i=1}^{d_s}p(\sbf^{(i)}_t|\zbf_{1:t}). 
\end{align*}

\paragraph{Full factorization}The full copula hierarchy is based on four primary dependencies: states and landmarks $c_{\thetabf(s,\ell)}$, landmarks and landmarks $c_{\thetabf(\ell,\ell)}$, and component dependency $c_{\thetabf(s)}$ and $c_{\thetabf(\ell)}$. The complete structure is given by the product
\begin{align}\label{eq:slam_copula}
	 c_{\thetabf}(\tilde{\sbf}_t,\tilde{\ellbf}_{1:M_t}) &= c_{\thetabf}^{(1)}(\tilde{\sbf}_t,\tilde{\ellbf}_{1:M_t})\cdot c_{\thetabf}^{(2)}(\tilde{\sbf}_t,\tilde{\ellbf}_{1:M_t}),\\
	 c_{\thetabf}^{(1)}(\tilde{\sbf}_t,\tilde{\ellbf}_{1:M_t}) &= c_{\thetabf(s,\ell)}(\tilde{\sbf}_t,\tilde{\ellbf}_{1:M_t})c_{\thetabf(\ell,\ell)}(\tilde{\ellbf}_{1:M_t}),\nonumber\\
	 c_{\thetabf}^{(2)}(\tilde{\sbf}_t,\tilde{\ellbf}_{1:M_t}) &= c_{\thetabf(s)}(\tilde{\sbf}_t^{(1:d_s)})
	 \prod_{j=1}^{M_t} c_{\thetabf(\ell)}(\tilde{\ellbf}_{j}^{(1:d_\ell)}).\nonumber
\end{align} 

Given the dependency structure \eqref{eq:slam_copula} and a fully-factorized distribution of univariate marginals, our new model class is defined as the set of all distributions $p(\sbf_t,\ellbf_{1:M_t}|\zbf_{1:t}) =$
\begin{align}\label{eq:slam_posterior_copula}
	 c_{\thetabf}(\tilde{\sbf}_t,\tilde{\ellbf}_{1:M_t})\prod_{i=1}^{d_s}p(\sbf^{(i)}_t|\zbf_{1:t}) \prod_{j=1}^{M_t}\biggl[\prod_{k=1}^{d_\ell}p(\ellbf^{(k)}_{j}|\zbf_{1:t})\biggr]. 
\end{align}

\section{Variational Copula SMC}\label{sec:vcslam}
In this section we introduce a variational SMC method that solves the SLAM problem using proposal distributions of the form \eqref{eq:slam_posterior_copula}. Each proposal $q(\xbf_t|\xbf_{1:t-1};\lambdabf)$ replaces $p(\cdot|\zbf_{1:t})$ and is indexed by the variational parameters $\lambdabf = \{\thetabf,\etabf\}$, representing the copula parameters $\thetabf$ and the marginal parameters $\etabf$. In what follows, we describe how to sample from the proposals and how to learn their variational parameters through gradient-based optimization.  

\subsection{A variational objective for fitting proposal distributions}
Our method fits proposal parameters to a set of observations using the Variational SMC \cite{smc:2018:naesseth_etal:variational_sequential_monte_carlo} framework. The objective is to minimize the Kullback Leibler (KL) divergence between the expected SMC approximation $\Ebf[q(\xbf_{1:T};\lambdabf)]$ and the target distribution $p(\xbf_{1:T}|\zbf_{1:T})$. This objective is intractable, but it can be upper-bounded by
\begin{align*}
	\mathsf{KL}(\Ebf[\ q(\xbf_{1:T};\lambdabf)] \ || \ p(\xbf_{1:T}|\zbf_{1:T})\ ] ) \leq -\Ebf\left[\log\frac{\hat{Z}_t}{Z_t}\right].
\end{align*}
The expectation here is taken with respect to the joint distribution of all variables generated by the SMC sampling procedure: $\phi(\xbf_{1:T}^{[1:N]},a_{1:T}^{[1:N]};\lambdabf) =$
\begin{align*}
	 \prod_{n=1}^N q(\xbf_1^{[n]};\lambdabf)\cdot\prod_{t=2}^T\prod_{n=1}^N  \rho_{t-1}^{[n]} q(\xbf_t|\xbf_{1:t-1}^{a_{t-1}^{[n]}};\lambdabf) \cdot \rho_T^{[n]},
\end{align*}
where $\rho_t^{[n]}= \tilde{w}^{a^{[n]}_{t}}_{t} / \sum_{j}\tilde{w}^{a^{[j]}_{t}}_{t}$. We describe this sampling procedure in Alg. \ref{alg:smc}. The log-normalization constant is
\begin{align}\label{eq:elbo_ratio}
	\log\hat{Z}_t = \sum_{t=1}^T\log\left( \frac{1}{N}\sum_{n=1}^N\tilde{w}_t(\xbf_{1:t}^{[n]})\right).
\end{align}

The dependence on the data and parameters enters the upper bound implicitly through the weights, proposed samples, and resampling procedure. Because $Z_T$ does not depend on the variational parameters, it can be treated a constant in the optimization. Therefore, minimizing the upper bound \eqref{eq:elbo_ratio} is equivalent to maximizing
\begin{align}\label{eq:elbo}
	\Ebf[\log \hat{Z}_T]
	&=\sum_{t=1}^T\Ebf_{\phi_t}\left[\log\left( \frac{1}{N}\sum_{n=1}^N \tilde{w}_t(\xbf_{1:t}^{[n]})\right) \right].
\end{align}
In contrast to previous adaptive SMC methods that minimize the KL of the target directly with the proposal \cite{smc:2015:gu_etal:neural_smc}, this approach optimizes the fit of the final SMC distribution to the true target.

\subsection{Computing gradients}
We assume the proposals can be reparameterized such that sampling $\xbf_t\sim q(\xbf_t|\xbf_{1:t-1};\lambdabf)$ is equivalent to evaluating the deterministic function $\xbf_t=f(\varepsilon_t,\xbf_{1:t-1};\lambdabf)$ with a sample $\varepsilon_t\sim q(\varepsilon;\lambdabf)$ \cite{vi:2013:kingma_welling:autoencoding_variational_bayes}. We then update the variational parameters using stochastic gradient steps, where the gradient is 
\begin{align*}
	\gbf_{\lambdabf} &= \nabla_{\lambdabf}\Ebf[\log \hat{Z}_T],\\
	&= \underbrace{\Ebf_{q(\varepsilon)}[\log \hat{Z}_T \nabla_{\lambdabf}\log q(\varepsilon;\lambdabf)]}_{\gbf_{\text{score}}} + \underbrace{\Ebf_{q(\varepsilon)}[\nabla_{\lambdabf}\log \hat{Z}_T]}_{\gbf_{\text{rep}}}.
\end{align*}
Here, $\gbf_{\text{score}}$ is the score function gradient and $\gbf_{\text{rep}}$ is the standard reparameterization gradient \cite{vi:2016:ruiz_etal:the_generalized_reparameterization_gradient}. To keep the optimization tractable, we approximate $\gbf_{\lambdabf} \approx \gbf_{\text{rep}}$. Similar to others \cite{smc:2018:naesseth_etal:variational_sequential_monte_carlo}, we found this to work well in practice. The reparameterization gradient is derived as
\begin{align}
	\gbf_{\text{rep}} &=\sum_{t=1}^T\Ebf_{\phi_t}\left[\nabla_{\lambdabf}\log\left( \frac{1}{N}\sum_{n=1}^N \tilde{w}_t(\xbf_{1:t}^{[n]})\right) \right],\nonumber\\
	&= \sum_{t=1}^T\Ebf_{\phi_t}\left[\sum_{n=1}^N w_t(\xbf_{1:t}^{[n]}) \nabla_{\lambdabf}\log\tilde{w}_t(\xbf_{1:t}^{[n]})\right],\nonumber\\
	&\approx \hat{\gbf}_{\text{VSMC}}^{(k)}= \sum_{t=1}^T\sum_{n=1}^N w_t(\xbf_{1:t}^{[n]}) \nabla_{\lambdabf}\log\tilde{w}_t(\xbf_{1:t}^{[n]}). \label{eq:gradient}
\end{align} 
A derivation \eqref{eq:gradient} is provided in Appendix \ref{sec:grep}. Given an iterate $\lambdabf_{k-1}$ we estimate the gradient by running SMC with our proposals $q(\xbf_t|\xbf_{1:t-1};\lambdabf)$ and evaluate $\hat{\gbf}_{\text{VSMC}}$. The iterate is then updated with $\lambdabf_{k} \gets \lambdabf_{k-1} +\alpha_k \hat{\gbf}_{\text{VSMC}}^{(k)}$, where the step sizes are scheduled to satisfy the Robbins Monroe conditions: $\sum_k\alpha_k =\infty$, $\sum_k\alpha_k^2 < \infty$.

The gradient $\nabla_{\lambdabf}\log\tilde{w}_t(\xbf_{1:t}^{(n)})$ will change depending on the variational parameter and whether we consider the marginal parameters or the copula parameters. We have
\begin{align*}
	\nabla_{\lambdabf}\log\tilde{w}_t(\xbf_{1:t}^{[n]}) &= \frac{\nabla_{\lambdabf}\tilde{w}_t(\xbf_{1:t}^{[n]})}{\tilde{w}_t(\xbf_{1:t}^{[n]})}, \\
	\nabla_{\lambdabf} \tilde{w}_t(\xbf_{1:t}^{[n]})&= \frac{\gamma(\xbf_{1:t}^{[n]})}{[q(\xbf_{t}^{[n]}|\xbf_{1:t-1}^{[n]};\lambdabf)]^2}\nabla_{\lambdabf}q(\xbf_{t}^{[n]}|\xbf_{1:t-1}^{[n]};\lambdabf).
\end{align*}
Under the reparameterization and the dependency model, the optimization naturally splits into two problems:
\begin{align*}
	\nabla_{\xbf}\log q(\xbf_{t}|\xbf_{1:t-1};\lambdabf) &= \nabla_{\xbf}\log c_{\thetabf}(\tilde{\xbf}_{1:d};\lambdabf) \\
	&+ \sum_{i=1}^d \nabla_{\xbf}\log q(\xbf_{i,t}| \xbf_{i,1:t-1};\lambdabf).
\end{align*}
When optimizing the copula parameters $\thetabf$, the second term disappears, and both terms contribute to the gradient by optimizing the marginals. This leads to the following expectation-maximization procedure, which we call VC-SMC (Alg. \ref{alg:vcslam}). 

VC-SMC holds the marginal parameters fixed while it optimizes the copula parameters. It then alternates: holding the copula parameters fixed to fit the marginals. The process repeats until the ELBO is maximized at convergence. After obtaining a good fit, we can sample from our proposals using Alg. \ref{alg:smc} to solve the SLAM problem. In practice, we compute the gradients with automatic differentiation, reducing the model requirements and permitting the algorithm to be implemented as a library.

\begin{algorithm}[H]
	\caption{VC-SMC}
	\label{alg:vcslam}
	\begin{algorithmic}[1]
		\State \textbf{input:} Observations $\zbf_{1:T}$, Model $p(\xbf_{1:T},\zbf_{1:T})$, Proposals $q(\xbf_t^{(i)}|\xbf_{1:t-1};\lambdabf)$, Copulas $c_{\thetabf(\cdot,\cdot)}$
		\While{$\Lcal(\thetabf,\etabf)$ has not converged}
			\State {\color{gray} \# Learn the dependency structure}
			\While{not converged}
				\State $\thetabf_{k+1} \gets \thetabf_k +\alpha_k \hat{\gbf}_{\text{VSMC},\thetabf}$
			\EndWhile
			\State {\color{gray} \# Learn the marginal structure}
			\While{not converged}
				\State $\etabf_{k+1} \gets \etabf_k +\alpha_k \hat{\gbf}_{\text{VSMC},\etabf}$
			\EndWhile
		\EndWhile
	\State {\color{gray} \# Sample from the VC-SMC posterior (Alg. \ref{alg:smc})}
	\State \textbf{output:} $\xbf_{1:T} \sim q(\xbf_{1:T}|\lambdabf)$
	\end{algorithmic}
\end{algorithm}

\begin{algorithm}[H]
	\caption{VC-SMC Posterior Sampling}
	\label{alg:smc}
	\begin{algorithmic}[1]
		\State \textbf{input:} Targets $p(\xbf_{1:t},\zbf_{1:t})$, Proposals $q(\xbf_t^{(i)}|\xbf_{1:t-1})$, Copulas $c_{\thetabf(\cdot,\cdot)}$, number of particles $N$
		\State {\color{gray} \# Initialize trajectory particles}
		\For{$n=1,\cdots, N$}
			\State $\xbf_1^{[n]} \sim q(\xbf_1;\lambdabf)$ 
			\State $\tilde{w}_1^{[n]} \gets \frac{ p(\xbf_1^{[n]})p(\zbf_1|\xbf_{1}^{[n]})}{q(\xbf_{1}^{[n]};\lambdabf)}$
		\EndFor
		\For{$t=1,\cdots,T$}
			\For{$n=1,\cdots, N$}
				\State $a_{t-1}^{[n]} \sim \text{Categorical}(N,\rho^{[j]}_{t-1})$
				\State $\xbf_t^{[n]} \sim q(\xbf_t|\xbf_{1:t-1}^{(a_{t-1}^{[n]})})$ 
				\State $\tilde{w}_t^{[n]} \gets \frac{ p(\xbf_t^{[n]}|\xbf_{1:t-1}^{[n]})p(\zbf_t|\xbf_{1:t}^{[n]},\zbf_{1:t-1})}{q(\xbf_{t}^{[n]}|\xbf_{1:t-1}^{[n]},\zbf_{1:t})}$
			\EndFor
		\EndFor
	\State $a_T \sim \text{Categorical}(N,\rho^{[j]}_{T})$
	\State \textbf{output:} $x_{1:T}^{a_T}$
	\end{algorithmic}
\end{algorithm}

\begin{algorithm}[H]
	\caption{VC-SMC Proposal Sampling}
	\label{alg:vcslam_proposal}
	\begin{algorithmic}[1]
		\State \textbf{input:} $\varepsilon$, $\lambdabf = \{\thetabf,\etabf\}$, 
		\State {\color{gray} \# Sample from the reparameterized function}
		\State $\xbf^{(i)}_t = f( \varepsilon, \xbf_{1:t-1};\etabf)$ $\forall$ $i =1,\cdots,d$
		\State \textbf{output:} $\xbf_{t} = \xbf^{(1:d)}_t$
	\end{algorithmic}
\end{algorithm}

\section{Related Work}
\subsection{Variational Inference}
Our variational algorithm (VC-SMC) is an extension of variational sequential Monte Carlo \cite{smc:2018:naesseth_etal:variational_sequential_monte_carlo} to support the class of copula-factorized proposal distributions. Others have also considered hierarchical copula models \cite{vi:2015:tran:copula_variational_inference,vi:Han2015VariationalGC} in a variational setting to preserve multivariate dependence between latent variables. Normalizing flows are another related set of inference methods, where the parameters of a nonlinear function mapping samples from a standard normal distribution to the posterior of interest are learned \cite{vi:rezende2015variational}. More related methods fall in the area of \emph{optimal transport} \cite{el2012bayesian, pulido2019sequential}, where again, parameters of a transformation between a `simple' distribution and the posterior of interest are obtained. The application of these methods to non-Gaussian SLAM problems has yet to be realized and could serve as an interesting area for future work.


\subsection{Non-Gaussian SLAM}
The broad focus of non-Gaussian SLAM research has largely been on nonparametric methods \cite{slam:2007:grisetti_etal:rao_blackwell_pf, slam:fourie2016nonparametric, slam:montemerlo2002fastslam, slam:fourie2017multi, eliazar2004dp}. Similar to our algorithm, these methods make no assumptions about the state or landmark distributions. However, their model classes are not able to separate the joint dependency from the marginal distributions in a way that permits decoupled modeling and inference. For example, the sum of Gaussians method \cite{slam:2003:durrant_whyte_etal:a_bayesian_approach} is a filtering-based approach that uses mixtures of coupled dependency and marginal behavior. The proposed copula decomposition enables the development of inference algorithms that flexibly incorporate a variety of assumptions about the marginals or dependency structure of the joint distribution, bridging the gap between fully parametric methods and nonparametric methods.

\section{Experiments}
In this section, we validate the main claims of our paper using experimental data gathered in simulation. We test if VC-SMC can improve inference accuracy in SLAM settings where uncertainty is non-Gaussian, and nonlinearities induce complex relationships between latent variables. Each experiment uses the Bootstrap Particle Filter (BPF) as its baseline \eqref{eq:bootstrap}. In each experiment, VC-SMC is trained using the Adam optimizer \cite{opt:2015:kingma_ba:adam_a_method_for_stochastic_optimization}, with a fixed learning rate of $\alpha = 10^{-2}$. More details about each experiment are provided in Appendix \ref{app:experiments}.

\subsection{Uncertain Data Association}
\label{sec:results:threedoors}
\begin{figure}
	\centering
	\includegraphics[width=\columnwidth]{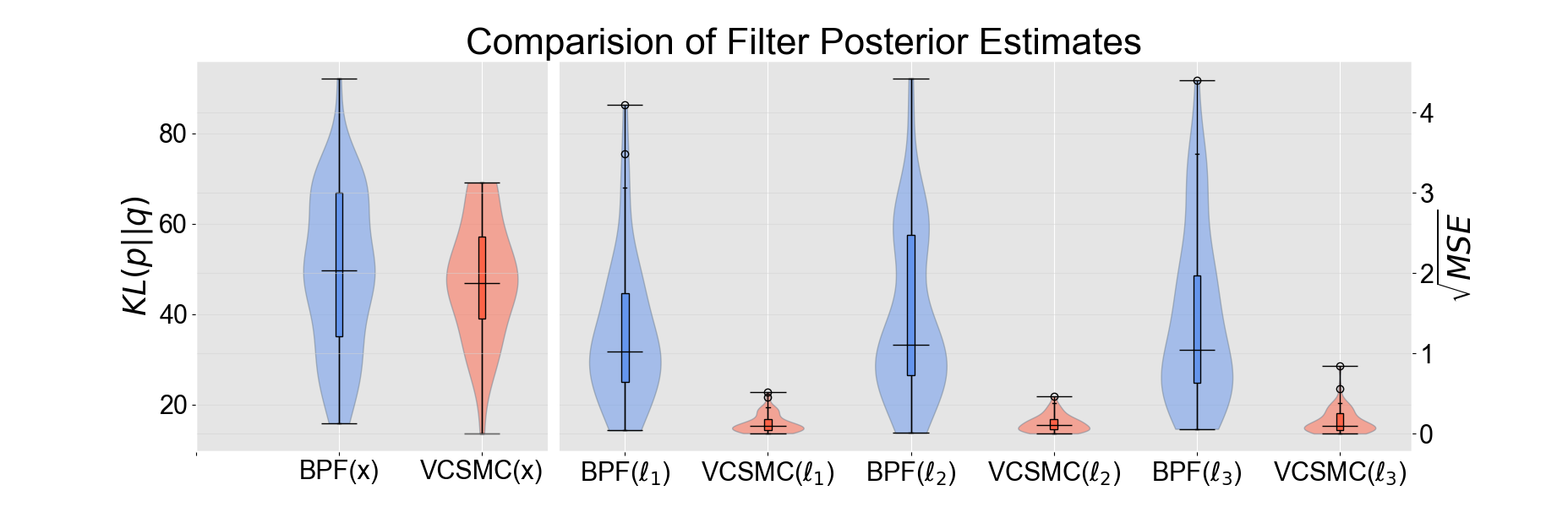}
	\caption{\textbf{VC-SMC can handle uncertain data association:} We show the KL divergence between the true posterior and pose distributions, and the root mean-squared error between ground truth landmark values and the posterior mean estimates of VC-SMC and the BPF. Data was gathered over 50 independent trials using 100 particles.}
	\label{fig:three_doors_comparision}
\end{figure}
\begin{figure}
	\centering
	\includegraphics[width=\columnwidth]{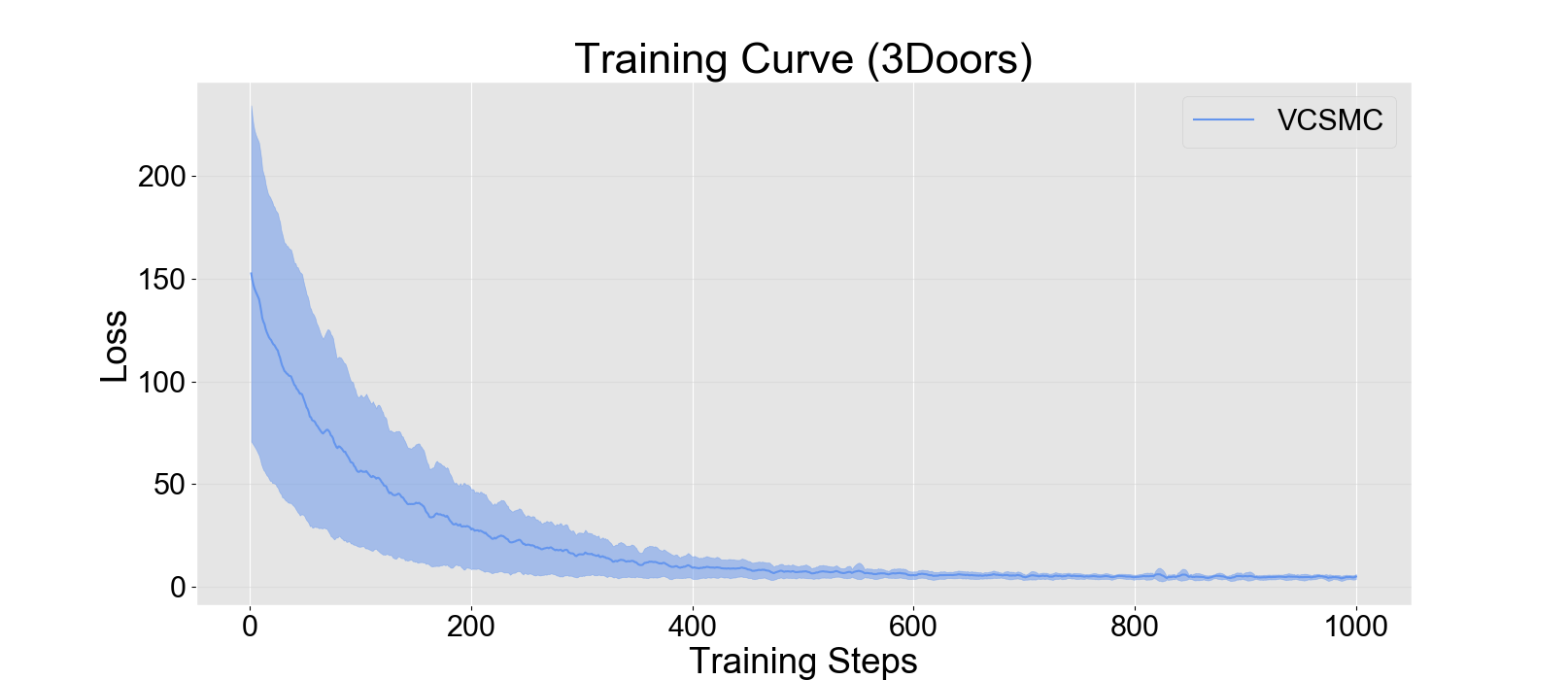}
	\caption{\textbf{Training progress in the 3Doors problem:} The variational loss \eqref{eq:elbo} converges from randomly-initialized parameters at a reasonable rate: in approximately 400 out of the 1000 total training iterations.}
	\label{fig:three_doors_elbo}
\end{figure}
\begin{figure*}
    \centering
        \begin{subfigure}{0.3\textwidth}
        \includegraphics[width=\columnwidth]{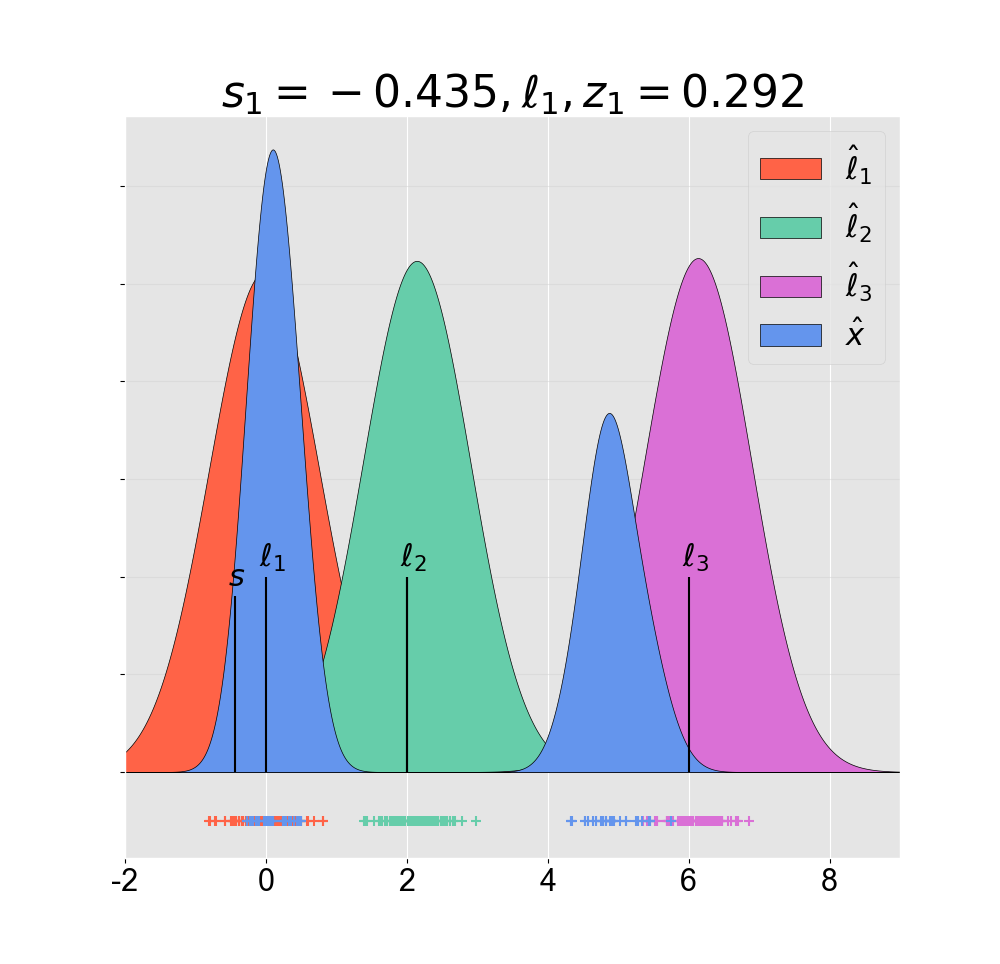}
        \caption{\textsc{vc-smc} first step}
    \end{subfigure}
        \begin{subfigure}{0.3\textwidth}
        \includegraphics[width=\columnwidth]{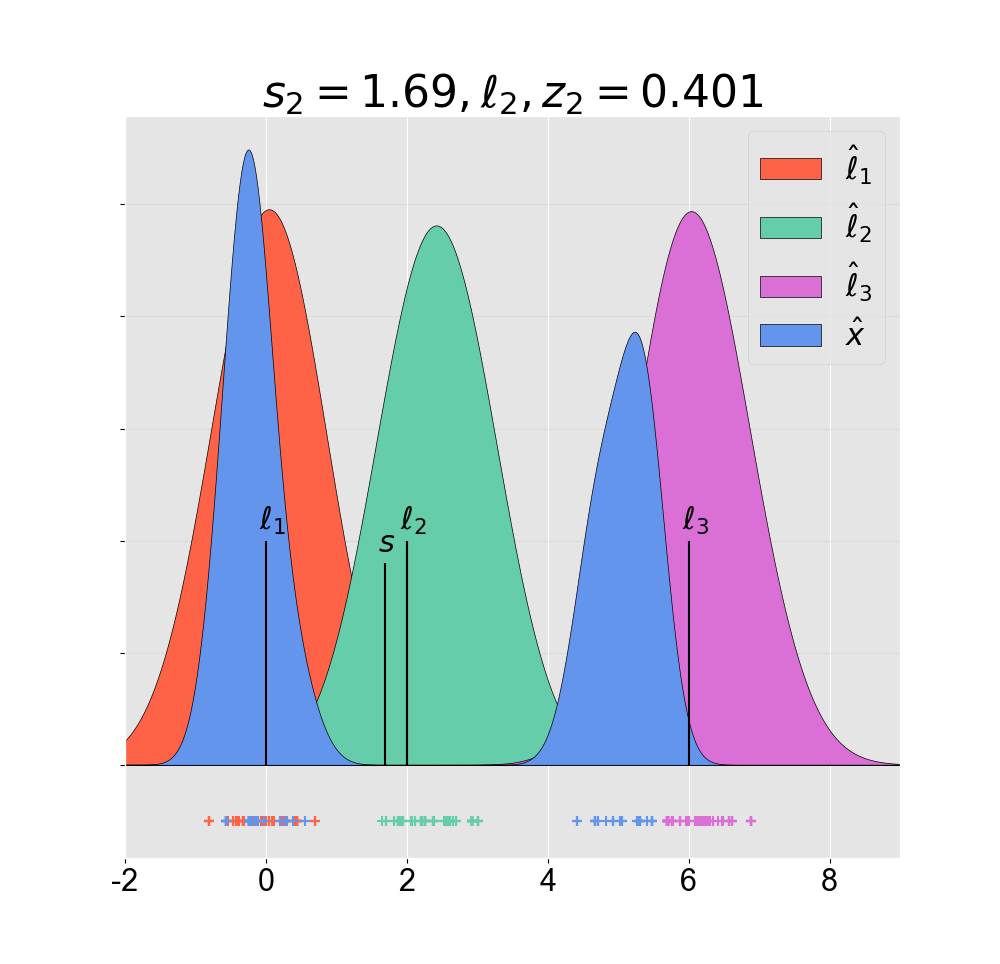}
        \caption{\textsc{vc-smc} second step}
    \end{subfigure}
    \begin{subfigure}{0.3\textwidth}
        \includegraphics[width=\columnwidth]{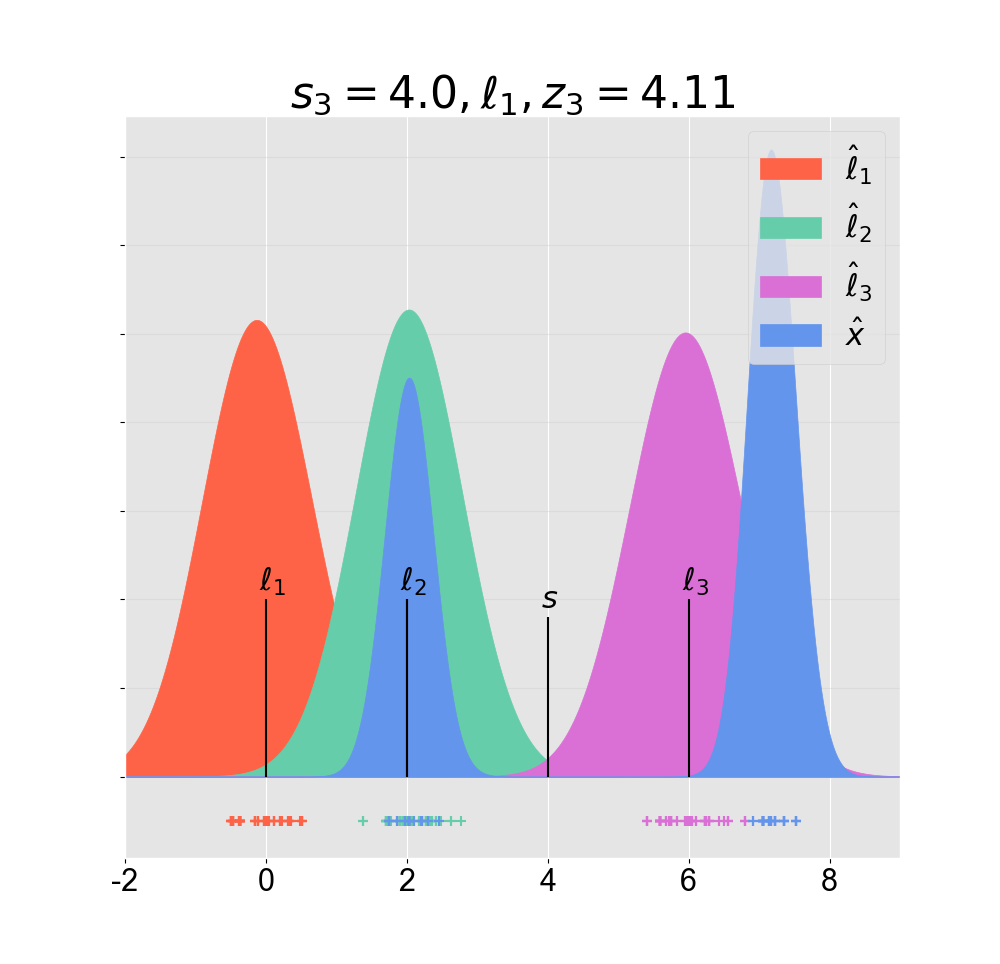}
        \caption{\textsc{vc-smc} third step}
    \end{subfigure}
    
    \begin{subfigure}{0.3\textwidth}
        \includegraphics[width=\columnwidth]{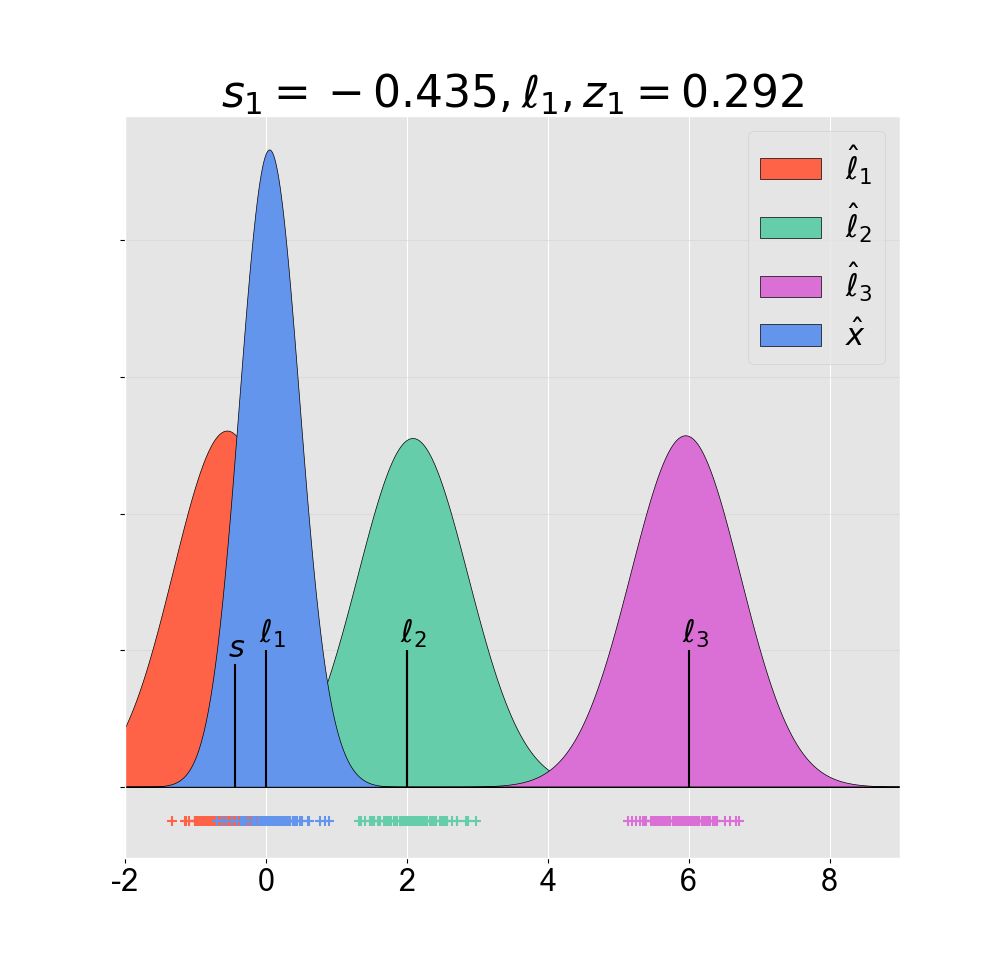}
        \caption{\textsc{bpf} first step}
    \end{subfigure}
    \begin{subfigure}{0.3\textwidth}
        \includegraphics[width=\columnwidth]{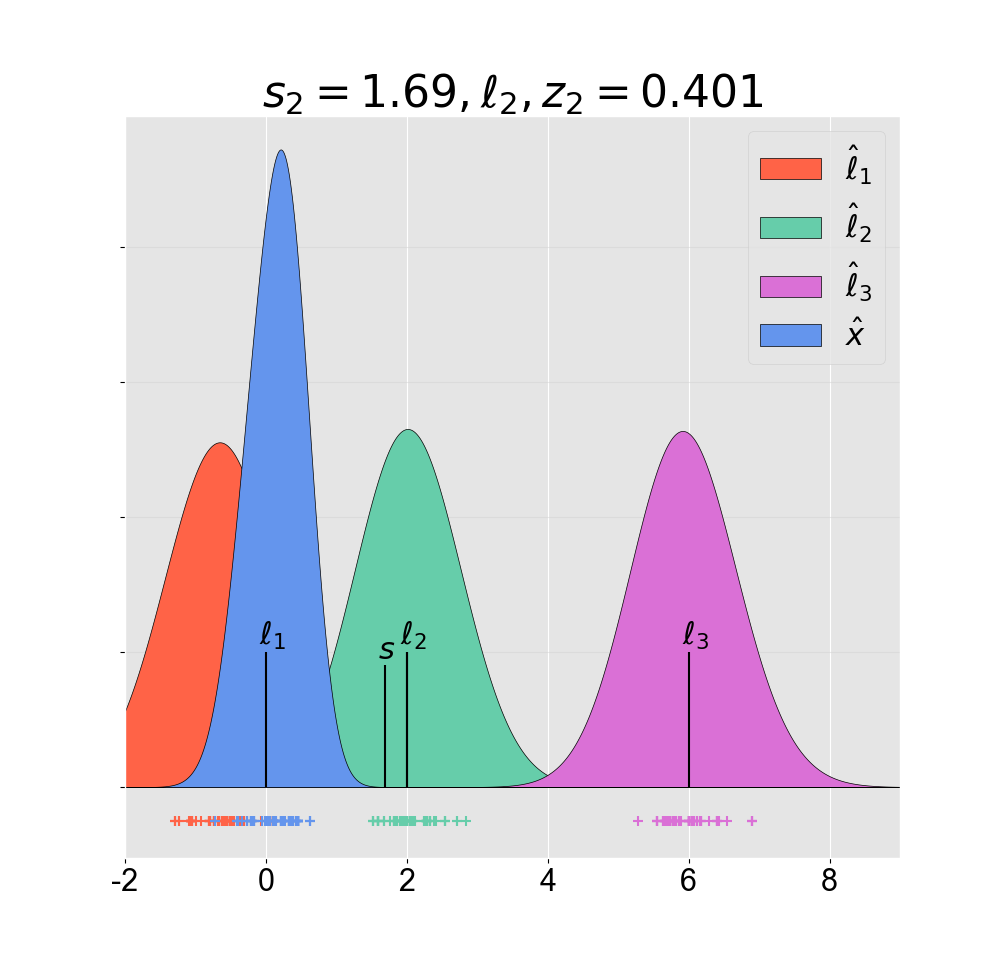}
        \caption{\textsc{bpf} second step}
    \end{subfigure}
    \begin{subfigure}{0.3\textwidth}
        \includegraphics[width=\columnwidth]{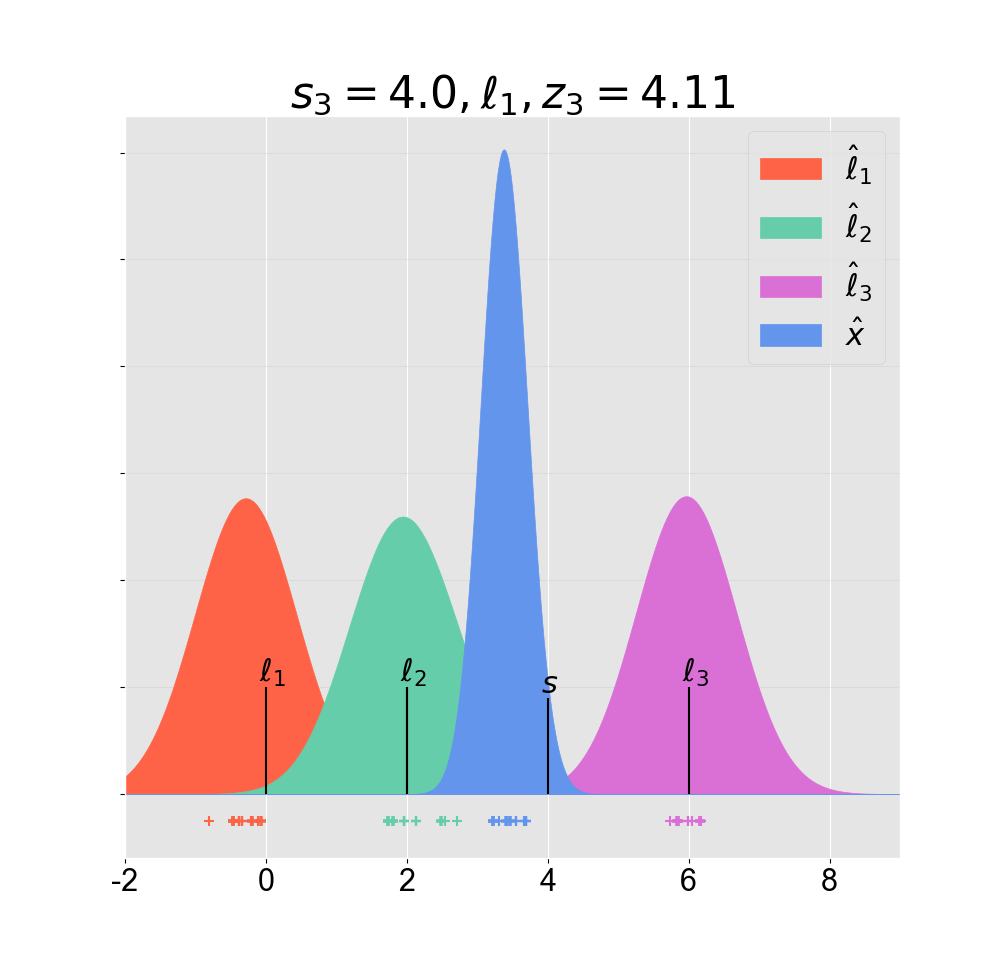}
        \caption{\textsc{bpf} third step}
    \end{subfigure}
    \caption{\textbf{Time evolution of the approximate target distribution in the 3Doors environment:} We show histograms of the three landmarks $\ell_i$ and robot pose at times $t=1,2,3$. The BPF fails to capture the multi-modal nature of the state distribution, and under-represents the posterior support. Each histogram is comprised of 500 particles.}
    \label{fig:threedoors-belief}
\end{figure*}
This experiment tests the hypothesis that VC-SMC can perform accurate inference in the presence of uncertain data association, where there is multi-modal, non-Gaussian uncertainty in the robot pose. The problem is set within the classic 3Doors environment \cite{slam:thrun2005probabilistic}. A robot moves along a single axis with constant velocity and observes the range to one of three doors serving as the navigation landmarks. Their absolute locations are $\ell_1=0$, $\ell_2=2$, $\ell_3=6$. 

This setting is challenging, because the robot does not know from which door its measurement is associated, and the number of possibilities compound at each time step. 
With Gaussian uncertainty in the transitions and observations, the exact posterior is a Gaussian mixture of three modes at $t=1$, nine modes at $t=2$, and twenty-seven modes at $t=3$.  To infer the latent variables and represent their uncertainty accurately, the robot must consider more than one hypothesis about its position and landmark locations.

We estimated the ground truth mean and variance parameters of the true posterior using a standard Bayes filter, and we compared the KL divergence between the pose distribution and estimates obtained with each test method. The root mean squared error was used for the unimodal landmark distributions. Distributions were parameterized with 100 particles. VC-SMC uses a three-component Gaussian mixture over the pose marginal and univariate Gaussians for the landmarks. These are linked through a Gaussian copula. The variational parameters of both methods were randomly initialized, and we trained them offline with 1000 gradient steps.
 
 Figure \ref{fig:three_doors_comparision} shows the distribution of root mean squared error, which was gathered from 100 independent trials. We can see that VC-SMC predictions have a significantly smaller error than the BPF method, which assumes a unimodal Gaussian proposal distribution. A shortcoming of the Gaussian proposal in the 3Doors environment is that it  provides inadequate coverage of the true posterior support -- assigning very little probability mass to regions of the latent space that potentially have high probability. In contrast, the VC-SMC proposal provides more modeling flexibility, which allows us to choose expressive models that provide better posterior coverage. Adjusting the mixture means is also straightforward through our variational algorithm.


In Fig. \ref{fig:threedoors-belief} we show qualitative results of the belief distribution at each step. We observe that even in this relatively simple problem BPF fails to accurately represent the full belief over robot poses and quickly loses modes of the posterior. This is due to the fact that the bootstrap proposal poorly covers the target distribution. In contrast, by optimizing the marginal and dependency parameters of non-Gaussian proposals VC-SMC is able to represent more modes in the posterior resulting from the data association uncertainty. In Figure \ref{fig:three_doors_elbo} we show the variational loss \eqref{eq:elbo} as it decreases throughout training. Based on the results of this section, we conclude that VC-SMC can support non-Gaussian inference in settings involving uncertain data association. 

\subsection{Nonlinear Planar Navigation}
\begin{figure}
    \centering
    \includegraphics[width=\columnwidth]{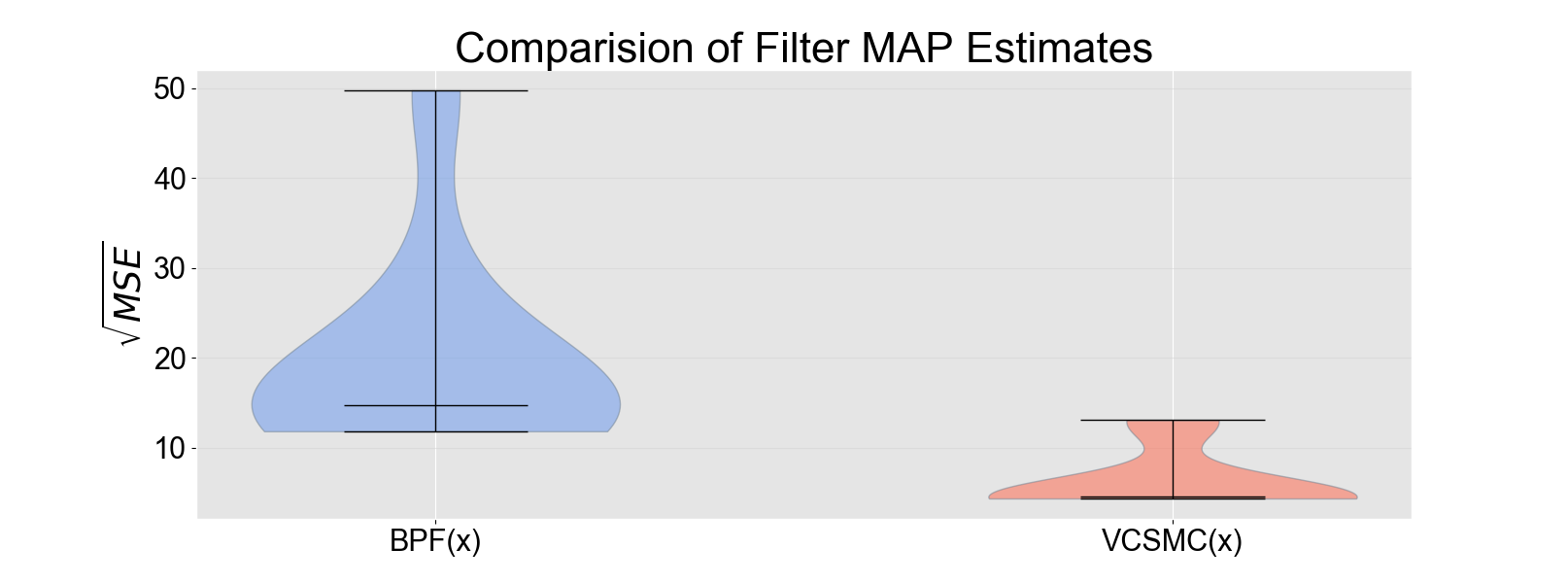}
    \caption{\textbf{Comparison of MAP estimates for Nonlinear Planar Navigation:} We compare the root mean-squared error of the predictions from BPF and VC-SMC. The variational method is able to achieve a superior fit since it can adjust the marginal and copula parameters. }
    \label{fig:nonlinear}
\end{figure}
\begin{figure}
	\centering
	\includegraphics[width=\columnwidth]{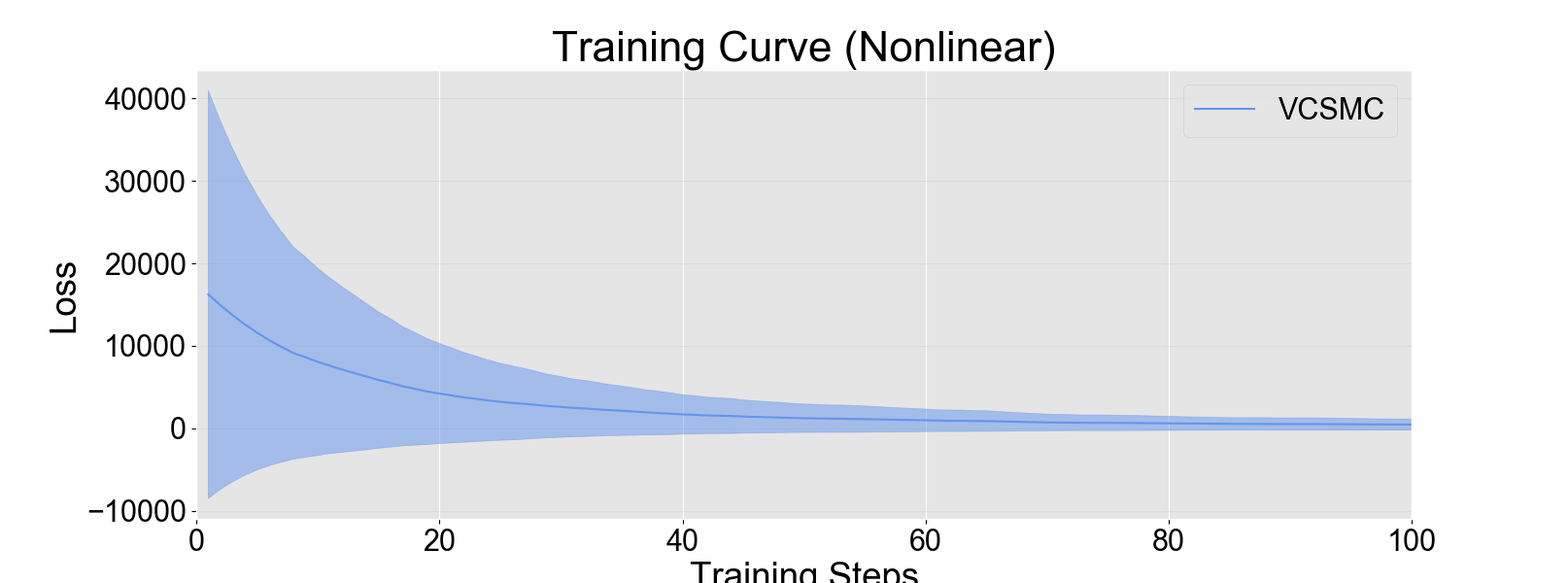}
	\caption{\textbf{Training progress in the Nonlinear Planar Navigation problem:} The variational loss \eqref{eq:elbo} converges from randomly-initialized parameters at a reasonable rate, in approximately 80 out of the 1000 total training iterations.}
	\label{fig:nonlinear_elbo}
\end{figure}

In this experiment we test the hypothesis that copula-factored distribution models can lead to improved inference accuracy in the presence of non-linearly distorted uncertainty. A robot wishes to localize its three-dimensional state over several time steps using measurements from the map origin, which are corrupted by additive Gaussian noise. States evolve according to the nonlinear transition model:
\begin{align*}
	x_{t+1} &= x_t + v \cos\theta_t, & y_{t+1} &= y_t + v \sin\theta_t, & \theta_{t+1} &= \theta_t.
\end{align*}
The robot moves with constant speed $v$ and heading $\theta$. This model has been used to simulate marine vehicles localizing with an ultra-short baseline acoustic system \cite{martin2017extending}.

In Fig. \ref{fig:nonlinear} we compare the total root mean-square error between the true reference trajectory and the estimates from both VC-SMC and BPF. We find that VC-SMC is able to achieve a lower error, since it is able to adjust the variational parameters at each time step, to fit the observed data. The plots represent data gathered over ten trials.

\section{Discussion}
This section concludes our paper with a brief discussion of the results, computational complexity, and possible avenues for future work.

\subsection{What have we learned?}
Starting from the observation that many SLAM problems are unrealistically approximated with Gaussian uncertainty, we asked if it was possible to improve predictive accuracy with more expressive models. To answer this, we introduced a new class of copula-factorized distributions that can represent any distribution with the added flexibility of separating multivariate dependency and marginal uncertainty into independent models. Indeed, these distributions were found to better represent domain uncertainty and lead to more accurate predictions compared to standard approaches when the model parameters were fit to maximize a lower bound to the log likelihood. 

\subsection{Computational Complexity} The per-iteration complexity of Algorithm \ref{alg:vcslam} is dictated by the total number of parameters $|\lambdabf|$. With $N$ particles evolving over $T$ time steps, a parameter update requires $\Ocal(TN)$ gradient computations with respect to each component in $\lambdabf$, as we show in Appendix \ref{sec:grep}. Each call of Algorithm \ref{alg:vcslam} computes $\Ocal(T N |\lambdabf|)$ partial derivatives. Since the number of parameters $|\lambdabf|$ \emph{per time-step} is roughly constant in typical applications, and the number of particles $N$ is fixed a priori, we expect that the per-iteration computational complexity of Algorithm \ref{alg:vcslam} is linear in time, in terms of the number of gradient computations. 

\subsection{Future work}
While the VC-SMC procedure permits inference with arbitrary copula models and non-Gaussian univariate marginals, the modeling freedom of the copula factorization lends itself to many new avenues for future work. Specifically, the factorization permits separate assumptions to be made about the joint dependency parameters than are made about the univariate marginal distributions. This is in contrast to assumptions of full joint Gaussianity. In this work, a particle-based inference method was chosen to preserve as much modeling flexibility as possible, but this came at the expense of efficiency. By developing algorithms tailored to a specific subset of copula models (for example, by considering specifically Gaussian copula models) or selection of univariate marginals, more efficient procedures for inference may be developed. Nonparanormal belief propagation \cite{elidan2012nonparanormal} is one example of such an existing method. Furthermore, leveraging data structures like the Bayes tree \cite{slam:kaess2010bayes} is of interest for improving the scalability of copula-based approaches.

\subsection{Conclusion}
In this work we sought to move beyond the Gaussian paradigm, to improve filtering accuracy in settings where states and landmarks have complex probabilistic dependence. We were able to exploit a powerful result from probability theory that factorizes joint relationships into marginal distributions and dependency models, called copulas. This copula factorization was used in a variational SMC filter (VC-SMC), where samples from the SLAM posterior were approximated, and parameters were fit through gradient based optimization. Our experimental results support the hypothesis that VC-SMC indeed improves inference accuracy when there is uncertain data association and when there was nonlinear propagation of uncertainty. We believe our results highlight the importance of modeling multivariate dependency for SLAM, as well as exemplify the benefits that copulas can offer the SLAM paradigm as a whole.

\section*{ACKNOWLEDGMENTS}
The authors would like to thank the anonymous reviewers for their helpful feedback. Additionally we would like to thank Qiangqiang Huang for his comments on an early draft.
This research was supported in part by the National Science Foundation, grant number IIS-1652064.

\bibliographystyle{IEEEtran}
\bibliography{ref}  


\newpage
\appendix
\subsection{Further Background on Non-Gaussian SLAM}
Particle filters have a long history of being applied to non-Gaussian SLAM problems. Gordon et al. \cite{gordon1993novel} introduced the bootstrap particle filter for non-Gaussian, bearing-only tracking problems. Bootstrapping was also used in FastSLAM method \cite{slam:montemerlo2002fastslam}, and this was shown to be effective for settings with ambiguous data association. Wang et al. \cite{wang2007upf} used an unscented particle filter and an unscented Kalman filter to improve estimation accuracy in comparison to FastSLAM. In an effort to improve proposal distributions, Marhamati et al. \cite{marhamati2012monte} used Monte Carlo approximations of the optimal proposal distribution.

There are a number of smoothing techniques for factor graphs with Gaussian measurement noise \cite{slam:2008:kaess_etal:isam, kaess2012isam2}. These methods generally make use of sparse nonlinear least-squares optimization. Within the scope of optimization-based approaches, a number of methods have been proposed to cope with multimodal noise, including max-mixtures \cite{olson2013inference} and the generalized prefilter method of Pfingsthorn et al. \cite{pfingsthorn2013simultaneous}. Gaussian smoothing methods have been extended by Fourie et al. \cite{slam:fourie2017multi} to the case of non-Gaussian inference using ``multimodal incremental smoothing and mapping'' (mm-iSAM), which has recently been demonstrated in the setting of object-level SLAM with ambiguous data associations \cite{slam:doherty2019multimodal}.

\subsection{Optimization of Gaussian Copula Parameters}
Optimization with respect to the Gaussian copula parameters $\thetabf$ is a constrained optimization problem. Specifically, a correlation matrix $\Pbf$ obtained from $\thetabf$ must be positive-definite and satisfy the constraint that the diagonal elements are all equal to 1, with all elements between -1 and 1. For matrices of size $M \times M$ satisfying these constraints, there are minimally $d = {M \choose 2}$ parameters. In order to construct a matrix that satisfies these properties, we first obtain a lower triangular Cholesky factor $\Lbf$, so that we may compute $\Pbf = \Lbf\Lbf^T$. Only some Cholesky factors $\Lbf$, however, will satisfy the diagonal constraint on $\Pbf$. Specifically, the Cholesky factor must be in the $M \times M$ \emph{oblique manifold} \cite{absil2009optimization}: 
\begin{align}
    \mathcal{OB}(M,M) &= \left\{ \Lbf \in \Rbb^{M\times M} : \textrm{diag}(\Lbf\Lbf^T) = \Ibf_M \right\}
\end{align}
In order to construct such a matrix from a set of parameters in $\Rbb^d$, we use the method of Lewandowski, Kurowicka, and Joe (the LKJ transform) \cite{lewandowski2009generating}, in which we fill the lower triangular portion of $\Lbf$ (excluding the diagonal) with the elements of $\thetabf \in \Rbb^d$. We then add the $M \times M$ identity matrix to $\Lbf$ to produce ones on the diagonal. Lastly, we normalize $\Lbf$ row-wise, which is necessary and sufficient for $\Pbf = \Lbf\Lbf^T$ to satisfy all of the constraints on correlation matrices. Furthermore, this transformation is differentiable with respect to the elements of $\thetabf$, which allows us to perform optimization of the dependency parameters in the \vcslam\ algorithm.

\subsection{Gradient of Log Evidence}\label{sec:grep}
 Here we derive the gradient of the log evidence with respect to the variational parameters \eqref{eq:gradient}:
 \begin{align*}
 	\gbf_{\text{rep}} &=\sum_{t=1}^T\Ebf_{\phi_t}\left[\nabla_{\lambdabf}\log\left( \frac{1}{N}\sum_{n=1}^N\tilde{w}_t(\xbf_{1:t}^{[n]})\right) \right],\\
 	&=\sum_{t=1}^T\Ebf_{\phi_t}\left[\nabla_{\lambdabf}\left\{ \log\left( \frac{1}{N} \right) + \log\left(\sum_{n=1}^N\tilde{w}_t(\xbf_{1:t}^{[n]})\right) \right\}\right] ,\\
 	&=\sum_{t=1}^T\Ebf_{\phi_t}\left[\nabla_{\lambdabf}\log\left(\sum_{n=1}^N\tilde{w}_t(\xbf_{1:t}^{[n]})\right)\right],\\
 	&= \sum_{t=1}^T\Ebf_{\phi_t}\left[\frac{1}{\sum_{j=1}^N\tilde{w}_t(\xbf_{1:t}^{[j]})}\sum_{n=1}^N \left(\frac{ \tilde{w}_t(\xbf_{1:t}^{[n]}) }{ \tilde{w}_t(\xbf_{1:t}^{[n]}) }\right)\nabla_{\lambdabf}\tilde{w}_t(\xbf_{1:t}^{[n]})\right] ,\\
 	&= \sum_{t=1}^T\Ebf_{\phi_t}\left[ \sum_{n=1}^N \left( \frac{\tilde{w}_t(\xbf_{1:t}^{[n]})}{\sum_{j=1}^N\tilde{w}_t(\xbf_{1:t}^{[j]})} \right) \nabla_{\lambdabf}\log\tilde{w}_t(\xbf_{1:t}^{[n]})\right],\\
 	&= \sum_{t=1}^T\Ebf_{\phi_t}\left[\sum_{n=1}^N w_t(\xbf_{1:t}^{[n]}) \nabla_{\lambdabf}\log\tilde{w}_t(\xbf_{1:t}^{[n]})\right],\\
 	&\approx \sum_{t=1}^T\sum_{n=1}^N w_t(\xbf_{1:t}^{[n]}) \nabla_{\lambdabf}\log\tilde{w}_t(\xbf_{1:t}^{[n]}) = \hat{\gbf}_{\text{VSMC}}^{(k)} 
 \end{align*} 
 
\subsection{Additional Experimental Details}\label{app:experiments}

\subsubsection{3Doors Domain}
In this problem a robot needs to estimate its position along a one-dimensional line using range observations obtained from one of three landmarks (i.e. doors), $\ell_1, \ell_2, \ell_3$. Uncertain data association will result from their observation across a temporal sequence of three vehicle poses $s_1, s_2, s_3$, which are all scalars. At each step the robot measures the relative distance $z_t\in\mathbb{R}^+$ to any one of three landmarks, and thus every landmark is equally likely to be detected. We model the associated observations from a three-component Gaussian mixture model as follows:
\begin{align*}
    p(z_t | s_t, \ell_{1:3}) = \sum_{i=1}^3c_i \cdot p(z_t | s_t, \ell_{i}).
\end{align*}
Here, the components $c_i$ are first sampled from a Categorical distribution, then from the respective landmark model 
\begin{align*}
  c_i &\sim \mathrm{Cat}([1/3, 1/3, 1/3]), & p(z | s, \ell_i) = \Ncal(\ell_i-s, \sigma_{z}^2).
\end{align*}
The state transition model is Gaussian distributed with a mean that evolves at constant velocity $\Delta=2$  
\begin{align*}
  p(s' | s) &= \mathcal{N}(s + \Delta, \sigma_{s}^2).
\end{align*}
The landmarks are assumed to be static between time steps with a small amount of Gaussian uncertainty:
\begin{align*}
  p(\ell' | \ell) &= \mathcal{N}(\ell, \sigma_{\ell}^2).
\end{align*}
We fixed the variances to be $\sigma^2_s = 0.1$ and  $\sigma^2_{\ell} = 0.1$.

\paragraph{Target distribution} The full joint distribution over three time steps is $p(\xbf_{1:3},z_{1:3})=$
\begin{align*}
     p(\xbf_1)p(z_1|\xbf_1)p(\xbf_2|\xbf_1,z_1)p(z_2|\xbf_2)p(\xbf_3|\xbf_2,z_{1:2})p(z_3|\xbf_3).
\end{align*}
The distributions at the first time step are given by 
\begin{align*}
    p(\xbf_1) = p(s_1)p(\ell_{1})p(\ell_{2})p(\ell_{3}).
\end{align*}
where $s_1 \sim \Ncal(0,\sigma_{s}^2)$, $\ell_1 \sim \Ncal(0, \sigma_{\ell}^2)$, $\ell_2 \sim \Ncal(2, \sigma_{\ell}^2)$, and $\ell_3 \sim \Ncal(6, \sigma_{\ell}^2)$.

\end{document}